\documentclass[letterpaper, 10 pt, conference]{ieeeconf}  

\IEEEoverridecommandlockouts                              
\usepackage[backend=biber,
            hyperref=true,
            url=false,
            isbn=false,
            doi=false,
            backref=false,
            style=ieee,
            natbib=true,
            mincitenames=1,
            maxcitenames=1,
            citestyle=numeric-comp,
            sorting=none,
            block=none]{biblatex}
        
\renewcommand{\bibfont}{\small}
\usepackage{flushend}
\usepackage{balance}

\usepackage{graphics} 
\usepackage{epsfig} 
\usepackage{amsmath} 
\usepackage{amssymb}  

\usepackage{algorithm}
\usepackage{algorithmicx, algpseudocode}
\usepackage{color}
\usepackage{soul} 
\let\labelindent\relax
\usepackage{enumitem}
\usepackage{xspace}

\newcommand{\etal}{et al.}

\newcommand{\eg}{\emph{e.g., }}

\newcommand{\figref}[1]{Figure\,\ref{fig:#1}}
\newcommand{\secref}[1]{Section\,\ref{sec:#1}}

\newcommand{\eqnref}[1]{{Eq.\ (\ref{eq:#1})}}
\renewcommand{\algref}[1]{{Algorithm\ \ref{alg:#1}}}

\title{\LARGE \bf
Continuous Relaxation of Symbolic Planner for \\ One-Shot Imitation Learning
}

\author{De-An Huang$^{1}$, Danfei Xu$^{1}$, Yuke Zhu$^{1}$, Animesh Garg$^{1,2}$, Silvio Savarese$^{1}$, Li Fei-Fei$^{1}$, Juan Carlos Niebles$^{1}$
\thanks{$^{1}$
All the authors
are with  with the Computer Science Department, Stanford University, 353 Serra Mall, Stanford, CA, USA. $^{2}$A. Garg is also with Nvidia, USA.
        {\tt\small \{dahuang,danfei,yukez,garg,feifeili,ssilvio,}
        {\tt\small jniebles\}@cs.stanford.edu}}%
}

\begin{document}

\maketitle
\thispagestyle{empty}
\pagestyle{empty}

\begin{abstract}
We address one-shot imitation learning, where the goal is to execute a previously unseen task based on a single demonstration. While there has been exciting progress in this direction, most of the approaches still require a few hundred tasks for meta-training, which limits the scalability of the approaches. 
Our main contribution is to formulate one-shot imitation learning as a symbolic planning problem along with the symbol grounding problem. This formulation disentangles the policy execution from the inter-task generalization and leads to better data efficiency. 
The key technical challenge is that the symbol grounding is prone to error with limited training data and leads to subsequent symbolic planning failures. We address this challenge by proposing a continuous relaxation of the discrete symbolic planner that directly plans on the probabilistic outputs of the symbol grounding model.
Our continuous relaxation of the planner can still leverage the information contained in the probabilistic symbol grounding and significantly improve over the baseline planner for the one-shot imitation learning tasks without using large training data.
\end{abstract}

\section{Introduction}

We are interested in robots that can learn a wide variety of tasks efficiently.
Recently, there has been an increasing interest in the \emph{one-shot imitation learning} problem~\cite{atkeson1997learning,duan2017one,finn2017one,huang2019neural,pathakICLR18zeroshot,xu2018neural,yu2018one}, where the goal is to learn to execute a previously unseen task from only a single demonstration of the task. 
This setting is also referred to as \emph{meta-learning}~\cite{vinyals2016matching,finn2017one}, where the meta-training stage uses a set of tasks in a given domain to simulate the one-shot testing scenario. This allows the learned model to generalize to previously unseen tasks with a single demonstration in the meta-testing stage.

The main shortcoming of these one-shot approaches is that they typically require a large amount of data for meta-training  (400 meta-training tasks in~\cite{huang2019neural} and 1000 in~\cite{xu2018neural} for the Block Stacking task~\cite{xu2018neural}) to generalize reliably to unseen tasks. However, this requirement is infeasible in realistic domains with hierarchical and long-horizon tasks that involve long-term environment interactions.

\begin{figure}[t]
  \centering
  \includegraphics[width=1.0\linewidth]{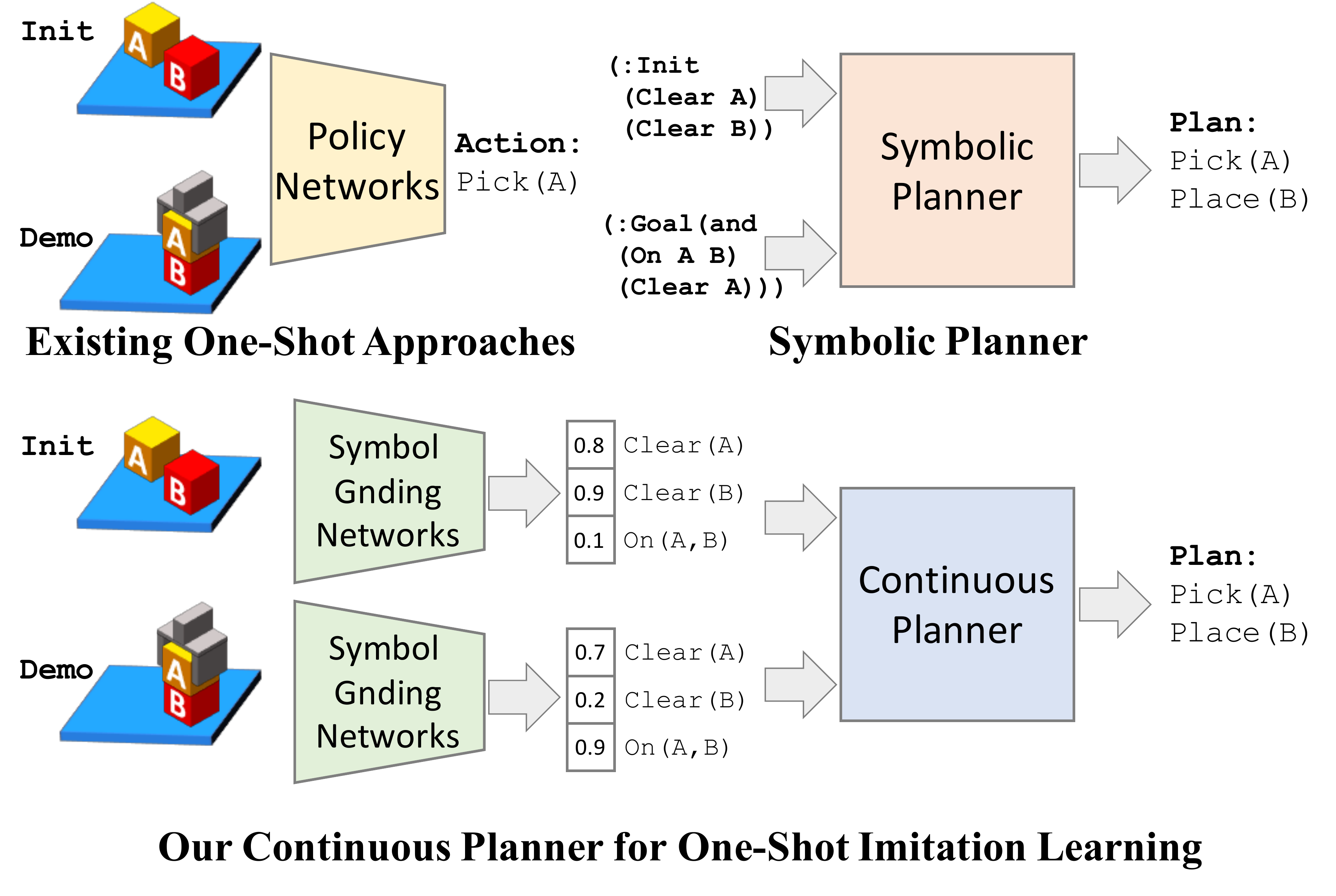}
  \vspace{-3mm}
  \caption{
  Our proposed Continuous Planner and Symbol Grounding Networks combine the advantages of existing one-shot imitation learning approaches with the symbolic planner. Compared to classical symbolic planners, our framework can take continuous states such as object locations and images as inputs. Compared to existing one-shot approaches, the use of symbolic planner in our model disentangles the inter-task generalization from policy execution and leads to much stronger generalization to unseen tasks.
  }
  \vspace{-3mm}
  \label{fig:fig1}
\end{figure}

The primary contribution of this paper is to formulate one-shot imitation learning for long-horizon tasks as a planning problem to leverage the structure of the symbolic domain definition.
This allows us to disentangle the policy execution from the inter-task generalization, which significantly reduces the number of tasks required in meta-training. In our formulation, the planner focuses on the policy execution, while the symbol grounding problem learns to map the continuous input state (\emph{e.g.,} object poses or images) to the  symbolic state representation required by the planner. In this case, the inter-task generalization is solely handled by the symbol grounding and disentangled from the policy execution. We argue that it is much easier for the symbol grounding problem to achieve inter-task generalization compared to the black-box policy networks used in previous works~\cite{duan2017one,huang2019neural,finn2017one,xu2018neural} because the symbol grounding function can be shared among the tasks in the same or similar domains. We further improve the generalization of symbol grounding by proposing a \emph{modular} Symbol Grounding Network (SGN) to infer the symbolic state of a given continuous input state. The modularity of our SGN enables effective parameter sharing among the symbolic states to further improve data efficiency.

The central technical challenge is that the outputs of the Symbol Grounding Network can be error-prone in low-data regimes. In this case, the SGN can output \emph{invalid} symbolic states that lead to subsequent symbolic planning
failures. 
Consider Block Stacking domain shown in \figref{fig1} as an example, it is possible (and likely as we will show in the experiments) for the SGN to output that both \texttt{On(A, B)} and \texttt{Clear(B)} are true. However, this is inconsistent, since both conditions cannot simultaneously be true.

Our solution to the challenge is to propose 
a continuous relaxation of the symbolic planner that replaces the set-theoretic representation~\cite{ghallab2004automated} in  the symbolic planner with the probabilistic symbols~\cite{konidaris2015symbol}.
This allows the planner to \emph{directly} plan on probabilistic distribution over the symbolic states.
This addresses the aforementioned inconsistency because the symbol grounding stage is no longer forced to make discrete decisions, and can simply provide continuous estimation of the symbolic state.
We show that our continuous relaxation of the symbolic planner can still leverage the information provided by the continuous outputs of the SGN to complete the task from a single demonstration. 

\figref{fig1} compares the proposed framework with the symbolic planner and other neural network based approaches for one-shot imitation learning. We refer to our continuous relaxation of the symbolic planner as Continuous Planner (CP). 
Compared to symbolic planners, our Continuous Planner can take continuous states as inputs. Compared to existing one-shot imitation models, our formulation decouples the demonstration interpretation model from the policy model and leads to better data efficiency.

In summary, the main contributions of our work are: (i) Formulating one-shot imitation as planning to disentangle the policy execution from inter-task generalization, leading to better data efficiency; (ii) Proposing Continuous Planner to allow the planner to directly operate on the distribution over the symbolic states and resolve the invalid state problem; (iii) Introducing modularity to the Symbol Grounding Neural Networks to further improve the inter-task generalization.

\section{Related Work}

\noindent \textbf{Structures in one-shot imitation learning.} The goal of one-shot imitation learning is to translate a task demonstration (observation) into an executable policy~\cite{duan2017one,finn2017one,goo2018one,xu2018neural,huang2019neural}. Real-world tasks are often long-horizon and multi-step, posing severe challenges for simple techniques such as behavior cloning-based methods~\cite{duan2017one}. Recent works in one-shot imitation learning aim to mitigate this challenge by imposing modular or hierarchical task structures in order to learn reusable subtask policies ~\cite{niekum2015learning,shiarlis2018taco,goo2018one,xu2018neural,huang2019neural}. 
NTP~\cite{xu2018neural} decomposes demonstration with hierarchical programs. NTG~\cite{huang2019neural} models the task structure as a graph generation problem. 
However, these works rely on function approximators to \emph{implicitly} model the transitions \textit{among} the subtasks, which often limits their ability to generalize. 
Our model explicitly \textit{grounds} the pre- and post-condition symbols of each subtask via our symbolic planning formulation. 

\vspace{1mm}
\noindent \textbf{High-level task planning}. Our planning formulation to handle the symbol uncertainty is primarily inspired by Konidaris \etal~\cite{konidaris2015symbol}, wherein probabilistic symbolic representation is used to replace the classic set-theoretic representation~\cite{ghallab2004automated} 
to express the symbol uncertainty in symbol acquisition. 
A major distinction in our formulation is that we still assume a \emph{deterministic domain}, where the transition between the symbolic states via actions is known and deterministic. 
This allows us to derive a continuous relaxation of the planner (\secref{CPDDL}) that greatly reduces the computational complexity. The relaxation assumes deterministic transitions while still handling uncertainty in symbol grounding. A similar deterministic assumption has also been leveraged under POMDP~\cite{littman1996algorithms,kaelbling1998planning,ross2008online}, in which the deterministic POMDP (DET-POMDP)~\cite{littman1996algorithms} formulation assumes deterministic transition function and known observation function. Our formulation can be viewed as a relaxed DET-POMDP formulation with an unknown observation function.
Such formulation involving uncertainty is also related to a large body of research in probabilistic planning~\cite{yoon2007ff,yoon2008probabilistic,little2007probabilistic,jetchev2013learning,huber2000hybrid}. However, these works only handle the uncertainty in the state transitions but not uncertainty in the symbols themselves~\cite{konidaris2015symbol}. Lastly, our work is also related to progress on integrating meta-learning and symbolic planning~\cite{chitnis2019learning,bhardwaj2017learning}. Chitnis \etal~\cite{chitnis2019learning} integrate ideas from task and motion planning with MAML~\cite{finn2017maml} to learn few-shot policies for continuous parameters used in planning.

\vspace{1mm}
\noindent \textbf{Symbol Grounding for Manipulation.}
Our formulation connects the symbolic planner with the continuous input state. 
This is related to progress on symbol grounding of geometric continuous state for manipulation planning~\cite{bidot2017geometric,gravot2005asymov,kaelbling2010hierarchical}. The main difference is that most of these approaches assume that they are given the mapping from geometric to symbolic states. On the other hand, our Symbol Grounding Networks does not assume a given mapping and can use arbitrary continuous states as input. The idea of learning this symbol grounding has also been explored for manipulation~\cite{abdo2012low,dearden2014manipulation,sjoo2011learning}. However, these approaches either do not consider planning on the symbol or do not take into account the symbol uncertainty in planning as our Continuous Planner.

\begin{figure*}[t]
  \centering
  \includegraphics[width=0.9\linewidth]{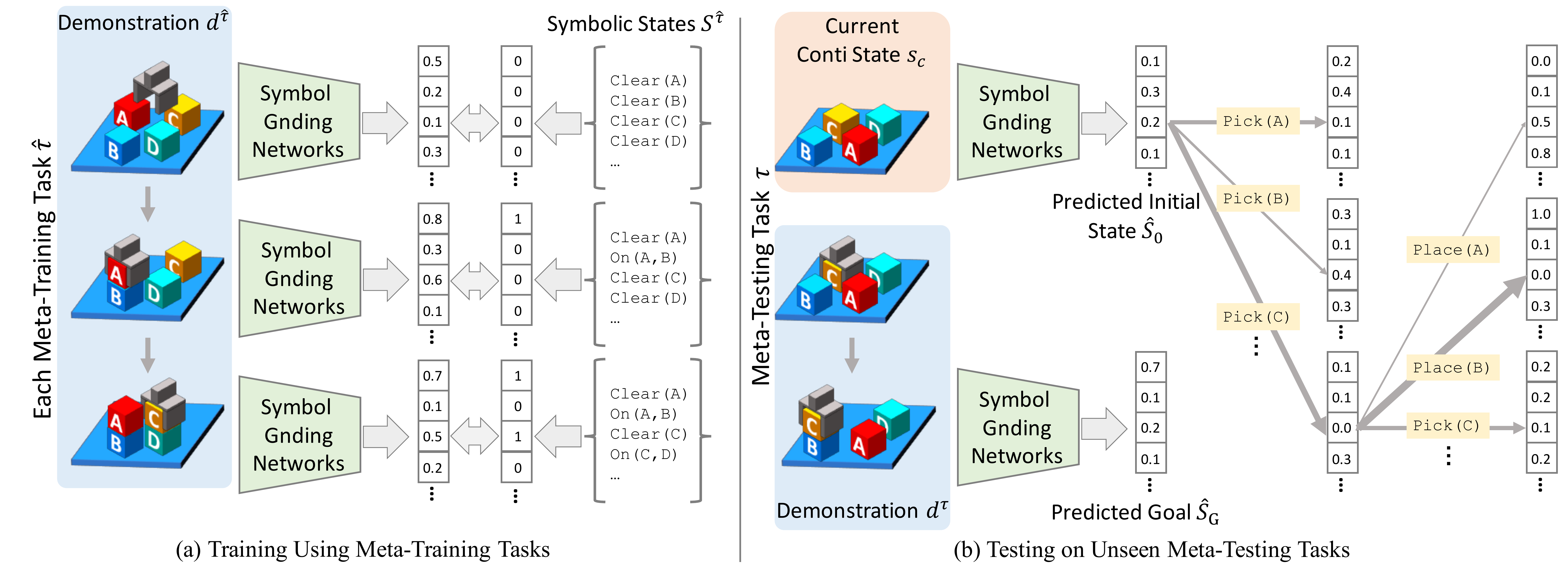}
  \vspace{-2mm}
  \caption{Overview of the training and testing of our framework. (a) We only need to train the Symbol Grounding Networks. We assume that the meta-training tasks contain the ground truth symbolic states aligned with the demonstrations. This is a valid assumption as the symbolic states can be automatically generated from the domain definition and the action annotation used in previous works~\cite{huang2019neural,xu2018neural}. (b) At test time, we apply our Symbol Grounding Networks to the demonstration to obtain the goal distribution and the current state distribution. We then use the Continuous Planner to generate plans. Note that the whole process can be repeated when the current continuous state changes. This allows our framework to work as a close-loop policy.
  }
  \vspace{-2mm}
  \label{fig:sys_fig}
\end{figure*}

\section{Problem Setup and Preliminaries}

\subsection{One-Shot Imitation Learning}
\label{sec:one_shot}

The goal of one-shot imitation learning~\cite{duan2017one} is to execute a previously \emph{unseen} task from a single demonstration.
Let $\mathcal{T}$ be the set of tasks in a domain, such as block stacking. Given a demonstration $d^{\tau} = [d^\tau_1, \dots, d^\tau_T]$ for task $\tau \in \mathcal{T}$, where each $d^\tau_t$ is a continuous state, the goal is to have a model $\phi(\cdot)$ to output a policy $\pi_\tau = \phi(d^\tau)$ to complete a new instance of the task. 
Modeling $\phi(\cdot)$ poses an extreme data efficiency challenge because it requires learning to interpret the demonstrations and to perform the task simultaneously. Most previous works adopt a meta-learning setup~\cite{finn2017maml}, where $\mathcal{T}$ is further divided into mutually exclusive sets of meta-training tasks $\mathcal{T}_{meta-train}$ and unseen meta-testing tasks $\mathcal{T}_{meta-test}$. $\phi(\cdot)$ is trained on $\hat{\tau} \in \mathcal{T}_{meta-train}$ so that $\pi_{\tau} = \phi(d^{\tau})$ can successfully complete the tasks $\tau \in \mathcal{T}_{meta-test}$. The assumption is that we have sufficient demonstrations and tasks in $\mathcal{T}_{meta-train}$ so the learned $\phi(\cdot)$ can generalize to tasks in $\mathcal{T}_{meta-test}$ at test time.

\subsection{Symbolic Planning}
\label{sec:pddl}

In this work, we formulate one-shot imitation learning as a classical symbolic planning problem. 
Following the definition in~\cite{ghallab2004automated}, a \emph{planning problem} $(S_0, S_G, O)$ contains an initial state $S_0$, a goal $S_G$, and a set of operators $O = \{o\}$.
Each operator is defined by:
$o = (name(o), precondition(o), effect(o))$,
where $name(o)$ includes the name of the operator and the list of arguments, $precondition(o)$ specifies the conditions that need to be satisfied in order to apply the operator, 
and $effect(o)$ 
defines how the state would be updated after applying the operator. 
The \emph{state} is defined as the set of all ground atoms that are true (\eg \{\texttt{On(A,B)}, \texttt{Clear(A)}\}). A ground atom is false if it is not presented in the state. A ground atom, such as \texttt{On(A,B)}, consists of the predicate (\texttt{On($\cdot$, $\cdot$)}) and the objects (\texttt{A} and \texttt{B}) for the arguments. 
An action $a$ is a grounded operator, where all the arguments of the operator are substituted by objects.
The solution of the planning problem is a plan $\Pi=[a_1, \dots, a_N]$ of a sequence of actions $a_i$. When starting at the initial state $S_0$, the plan will lead to a state that satisfies the goal $S_G$. 

We use the Planning Domain Definition Language (PDDL) to specify our planning problem. In PDDL, the planning problem is split into a \textit{domain} file and a \textit{problem} file. The domain file contains $O = \{o\}$ the set of operators and the predicates. The problem file contains the initial state $S_0$ and the goal $S_G$. 
Consider block stacking domain as an example. All block stacking tasks share the same domain file, while each of the tasks has its own problem file to specify the initial configuration/state and goal configuration/state. When we say  $\mathcal{T}_{unseen}$ and $\mathcal{T}_{seen}$ are in the same domain, we assume that they share the same domain file and is available at training.

\section{Our Method}

We address the one-shot imitation learning problem, where the goal is to complete a previously unseen task based on a single demonstration.
We formulate it as a planning problem and use our Symbol Grounding Networks (SGN) to explicitly ground the symbols. 
This disentangles the policy execution from the inter-task generalization and leads to better data efficiency compared to previous one-shot approaches. 
The key challenge is that the SGNs are error-prone with limited training data and can output ``invalid'' states. This leads to subsequent failure of the symbolic planner. We address this challenge by proposing the Continuous Planner to directly plan on the probabilistic outputs of SGNs. An overview of our approach is shown in \figref{sys_fig}. We will first formulate one-shot imitation learning as a planning problem in \secref{one_shot_plan}. Next, we will discuss the details of our Symbol Grounding Networks and Continuous Planner in \secref{SGN} and \secref{CPDDL}. Finally, we will include details of learning and inference of our framework in \secref{learning}.

\subsection{One-Shot Imitation as a Planning Problem}
\label{sec:one_shot_plan}

Existing one-shot imitation learning methods~\cite{duan2017one,huang2019neural,xu2018neural} parameterize $\phi(\cdot)$ as policy models conditioned on demonstrations. While these methods have been shown to generalize to $\mathcal{T}_{meta-test}$, training such policy networks requires a large amount of data in $\mathcal{T}_{meta-train}$ because the policy networks need to simultaneously interpret demonstrations and perform tasks. We formulate one-shot imitation as a symbolic planning problem: We disentangle the modeling of compound $\phi(\cdot)$ into learning a Symbol Grounding Networks (SGN) and perform Continuous Planning (CP): $\phi(\cdot) = CP(SGN(\cdot))$. In this case, the inter-task generalization is handled by the SGN, while the CP can focus on policy execution. Such disentanglement significantly reduces the complexity of generalizing to unseen tasks. 

Now we introduce the symbolic planning formulation of one-shot imitation learning. We can think of one-shot imitation learning as specifying the goal of the task using a demonstration. In this case, if we can map the demonstration to a symbolic goal $S_G$, and map the current observing continuous state $s_{c}$ to the corresponding symbolic state $S_0$, then we can use planning to solve the task based on the operators $O$ defined in the domain file. We observe that both $S_0$ and $S_G$ can be obtained by solving the symbol grounding problem that maps a continuous state $s$ to the corresponding symbolic state $S$. For $S_G$, we can use the symbol grounding of the demonstration $d^\tau=[d^\tau_1, \dots, d^\tau_T]$ for a task $\tau$. The symbolic state for the final continuous state $d^\tau_T$ is guaranteed to satisfy the goal. 
As for the initial state $S_0$, we can obtain it by symbol grounding of the currently observing continuous state $s_{c}$. We address symbol grounding by proposing the Symbol Grounding Networks (SGN). By predicting both the current and the goal symbolic states, we have:
\begin{equation}
\label{eq:decomp}
\Pi = CP(\hat{S_0}, \hat{S_G}, O) =CP(SGN(s_c), SGN(d^\tau_T), O). 
\end{equation}
Our formulation operates as a closed-loop policy conditioned on the demonstration if we view the plan  $\Pi$ as a multi-step action.
We can further improve $\hat{S}_G$ by conditioning the goal recognition on the entire demonstration $d^\tau$ instead of just the final observation $d^\tau_T$. 
The outline is shown in \figref{sys_fig}(b).

\subsection{Symbol Grounding Networks}
\label{sec:SGN}

\eqnref{decomp} formalizes the disentanglement of symbol grounding and planning. Based on this formulation, CP is independent of the inter-task distribution shift and focuses on the policy execution. In this case, the Symbol Grounding Networks (SGN) plays an important role to handle the inter-task generalization. Given a demonstration $d^\tau$ for an unseen task $\tau$,
the SGN needs to recognize the corresponding goal $S_G$ despite being trained only on meta-training tasks. More generally, the SGN needs to map the current continuous state $s_c$ from any task to the corresponding symbolic state. While still challenging, SGN is easier to optimize than the original compound problem because we can expect the symbol grounding to be shared among similar domains.

\begin{figure}[t]
  \centering
  \includegraphics[width=1.0\linewidth]{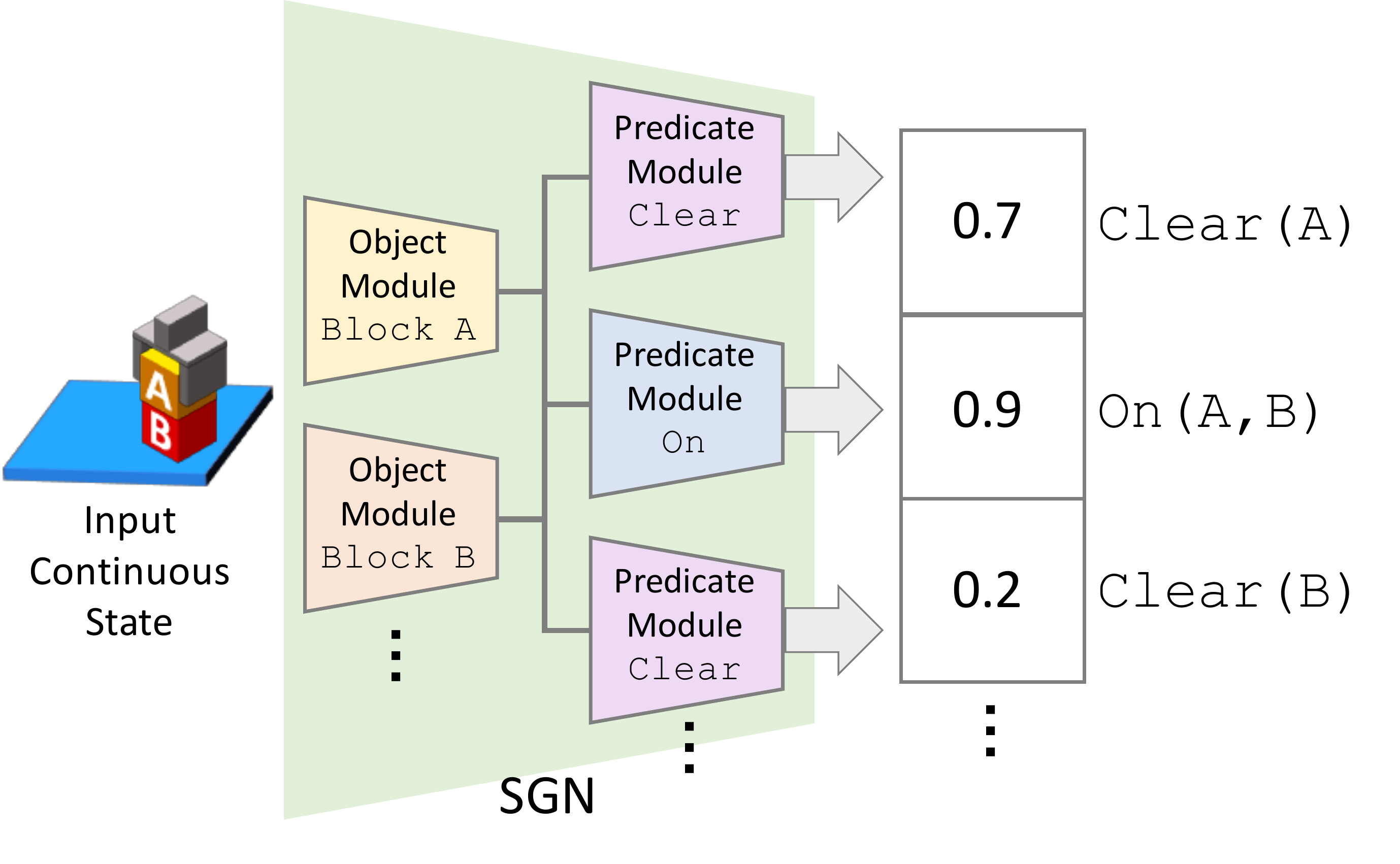}
  \vspace{-2mm}
  \caption{Example of three ground atoms with our modular Symbol Grounding Networks (SGN). \texttt{Clear(A)} and \texttt{Clear(B)} share the same predicate module $f_{\texttt{Clear}}$. The two ground atoms further share all the object modules with the ground atom \texttt{On(A, B)}. This parameter sharing using modularity allows us to improve the data efficiency of our SGN.
  }
  \vspace{-2mm}
  \label{fig:sgn}
\end{figure}

Nevertheless, as SGN plays an important role for the inter-task generalization of our model, we still need to make sure that it is data efficient. We thus leverage the recent progress on modular neural networks~\cite{andreas2016neural}, which has shown to improve generalization in Visual Question Answering~\cite{andreas2016neural} and Policy Learning~\cite{wang2018nervenet} by parameter sharing among model components. Let $s_c$ be the current continuous state input to the SGN. For each predicate \texttt{p}, we have a predicate module $f_\texttt{p}(\cdot)$, and for each object \texttt{b}, we have a object module $g_\texttt{b}(\cdot)$. Both are parameterized by Multi-Layer Perceptrons. A ground atom \texttt{p(b1,b2)} is classified by:
\begin{equation}
\label{eq:sgn}
    P(\texttt{p(b1,b2)}|s_c) = f_\texttt{p}(cat([g_\texttt{b1}(s_c), g_\texttt{b2}(s_c)])),
\end{equation}
where each object module $g_\texttt{bi}$ extracts an embedding from the current continuous state $s_c$. The embeddings from the object modules are later concatenated by $cat(\cdot)$ and fed as input to the predicate module $f_\texttt{p}$. This allows us to share the parameters for symbol grounding of each of the ground atom in the domain and improve the data efficiency of our SGN. \figref{sgn} shows examples of the ground atoms.

\subsection{Continuous Planner}
\label{sec:CPDDL}
We have discussed how to map a continuous state $s_c$ to a probabilistic symbolic state through our SGN in \eqnref{sgn}. 
The next step is to perform planning on the outputs. As the symbolic planner requires discrete symbolic state input, an obvious approach is to discretize the probabilistic output (\eg with a threshold 0.5). 
The main drawback of such strategy is that there is no guarantee that it would yield a ``valid'' symbolic state. For instance, after the discretization, it is possible that the output symbolic set would contain \texttt{On(A,B)} (\texttt{A} is on \texttt{B}) and \texttt{Clear(B)} (\texttt{B} has nothing on top), which cannot be both true at the same time.
Addressing such challenge is particularly crucial for one-shot imitation learning in low-sample regimes because the SGN is more likely to make inaccurate predictions.

Without additional handcrafted rules, invalid state checking can be posed as a satisfiability (SAT) problem given the planning domain, but applying it to every neural network output is computationally prohibitive and it is still non-trivial to map invalid states to valid ones.
We address this challenge by proposing a continuous relaxation of the symbolic planner to allow it to \emph{directly} plan on the probabilistic outputs in \eqnref{sgn}. 
We achieve this by replacing the set-theoretic representation~\cite{ghallab2004automated} in deterministic planner with probabilistic symbols~\cite{konidaris2015symbol} and aim to find a plan towards the recognized goal distribution. This allows our Continuous Planner to output a list of actions based on the SGN outputs.

Classical symbolic planning involves the following key steps: (i) Define the current symbolic state; (ii) Find the list of applicable actions; (iii) Select one of the applicable actions; (iv) Apply the action and arrive at a new state; and (v) Stop when reaching the goal. We now explain the continuous relaxation of all these five steps by replacing the symbolic states with probabilistic symbols or distributions over states. More importantly, because of the discrete and deterministic nature of the planning problem, we can derive efficient iterative formulas for all these steps based on the outputs of SGN without marginalizing a large state space.

\vspace{1mm}
\noindent\textbf{State Representation.} 
To handle the uncertainty in the outputs of SGN, we adopt the probabilistic symbolic states representation from~\cite{konidaris2015symbol}. For example, assume that we only have three possible ground atoms: \texttt{Clear(A)}, \texttt{Clear(B)}, \texttt{On(A, B)}. If the SGN outputs the following probabilities:
\begin{align}
     P(\texttt{Clear(A)}|s_c)&=0.6\\P(\texttt{Clear(B)}|s_c)&=0.3, \\P(\texttt{On(A, B)}|s_c)&=0.7,
\end{align}
then it can be seen as specifying a distribution over all the 8 possible symbolic states. We will use this probabilistic distribution as the state representation of our continuous planner instead of the set of true ground atoms.  We use $Z(s)$ to denote the distribution over symbolic states, which maps a symbolic state $s$ to its corresponding probability. Given that we know the set of all ground atoms, and assume conditional independence among the ground atoms, we can represent $Z(s)$ compactly with $P_{Z(s)}(g)$ the probability that ground atom $g$ is true given the distribution $Z(s)$ for all $g$.

\begin{algorithm}[t]
\caption{Training Procedure}
\label{alg:training}
\begin{algorithmic}
\State \textbf{Inputs:}  $\mathcal{D}_{meta-train}=\{(d^{\hat{\tau}}, a^{\hat{\tau}})\}$ containing demos $d^{\hat{\tau}}$ and actions $a^{\hat{\tau}}$ for ${\hat{\tau}} \in \mathcal{T}_{meta-train}$, domain operators $O$
\For {$(d^{\hat{\tau}}, a^{\hat{\tau}})\in \mathcal{D}_{meta-train}$}
    \State $S^{\hat{\tau}} \gets SymbolicStates(d^{\hat{\tau}}, a^{\hat{\tau}}, O)$ 
    \For {$(d^{\hat{\tau}}_t, S^{\hat{\tau}}_t) \in d^{\hat{\tau}}, S^{\hat{\tau}}$}
        \State $\hat{S}^{\hat{\tau}}_t \gets SGN(d^{\hat{\tau}}_t)$ \Comment apply symbol grounding
        \State $\mathcal{L} \gets CrossEntropy(\hat{S}^{\hat{\tau}}_t, S^{\hat{\tau}}_t)$
        \State $Adam(\mathcal{L}, SGN(\cdot))$  \Comment backprop and optimize
    \EndFor
\EndFor
\end{algorithmic}
\end{algorithm}

\vspace{1mm}
\noindent\textbf{Applicable Actions.} As discussed in \secref{pddl}, a symbolic action is applicable if all the ground atoms in its precondition are true. 
Now that our state representation is no longer discrete, 
there is no more ``applicable'' or ``inapplicable'' actions. Instead, 
given the current distribution  $Z(s)$,
the applicable probability of an action $a$ is:
\begin{equation}
\label{eq:a_app}
    \sum_{s, \gamma(s, a)} Z(s) = \prod_{g \in Pre(a)} P_{Z(s)}(g), 
\end{equation}
where $\gamma(s, a)$ means that $s$ satisfies the precondition set $Pre(a)$ of $a$. 
This summation can be represented by the probabilities of the ground atoms in the precondition set because of the conditional independence.

\vspace{1mm}
\noindent\textbf{Action Selection.} As defined in \eqnref{a_app}, we no longer have a list of applicable actions to apply, but the likelihood of the actions being applicable. In this case, the selection of actions becomes a ranking of the applicability of actions.

\vspace{1mm}
\noindent\textbf{Action Application.} In symbolic planners, the current symbolic state is moved to a new symbolic state by action application. We would like an analogous notion for distribution over states. In this case, we consider the shift of the distribution over states by an \emph{attempt} of the action. Given the current distribution over states, the distribution would shift if we attempt an action because it is possible that the action can succeed and change the current symbolic state. At the same time, it is also possible that the action may fail because the precondition is not satisfied. The new distribution $Z'(s')$ after attempting $a$ from $Z(s)$ can be described by:
\begin{equation}
    Z'(s') = \sum_{s, a(s) = s', \gamma(s, a)} Z(s) + \sum_{s, s = s', \neg \gamma(s, a)} Z(s),
\end{equation}
where $a(s)$ is the symbolic state we get by applying $a$ to $s$ and we use $\gamma(s, a)$ to represent that $s$ satisfies $Pre(a)$. The first term captures the transitions that the attempt of $a$ is successful and the second term captures the failure of the action.
Based on this definition, the probability of a specific ground atom in the new distribution $Z'(s')$ is:
\begin{align}
& P_{Z'(s')}(g) = \sum_{s', g\in s'} Z'(s') \\ 
&= \sum_{s, g\in a(s), \gamma(s, a)} Z(s) + \sum_{s, g \in s, \neg \gamma(s, a) } Z(s). \label{eq:app_g}
\end{align}
We consider three types of ground atom $g$: 
(i) $g$ is in the positive effect set of $a$. In this case, we know automatically that $g\in a(s)$ is true as long as we have $\gamma(s, a)$. In addition, we also have $g \not\in Pre(a)$. \eqnref{app_g}  can be rewritten as:
\begin{align}
&P_{Z'(s')}(g) = \sum_{s, \gamma(s, a)} Z(s) + \sum_{s, g \in s, \neg \gamma(s, a) } Z(s) \\
&=  \prod_{\hat{g} \in Pre(a)} P_{Z(s)}(\hat{g}) + (1 -  \prod_{\hat{g} \in Pre(a)} P_{Z(s)}(\hat{g})) P_{Z(s)}(g).
\end{align}
(ii) $g$ is in the negative effect set of $a$. In this case, we have $g\not\in a(s)$ when $\gamma(s, a)$, and the first term in \eqnref{app_g} is $0$. For the second term, we have $g \in Pre(a)$, and thus:
\begin{align}
    P_{Z'(s')}(g) &= \sum_{s, g \in s, \neg \gamma(s, a) } Z(s)\\
    &= P_{Z(s)}(g) - \prod_{\hat{g} \in Pre(a)} P_{Z(s)}(\hat{g}).
\end{align}
(iii) $g$ is unaffected by $a$. In this case, $g\in a(s)$ means that $g \in s$, and \eqnref{app_g}  can be rewritten as:
\begin{align}
P_{Z'(s')}(g) &= \sum_{s, g\in s, \gamma(s, a)} Z(s) + \sum_{s, g \in s, \neg \gamma(s, a) } Z(s)\\ &= \sum_{s, g\in s} Z(s) = P_{Z(s)}(g) .
\end{align}
These derivations allow us to efficiently compute the action application iteratively using $P_{Z(s)}(g)$ without marginalizing over the large discrete state space.

\setlength{\textfloatsep}{4mm}
\begin{algorithm}[t]
\caption{Testing Procedure}
\label{alg:testing}
\begin{algorithmic}
\State \textbf{Inputs:}  $\mathcal{D}_{meta-test}=\{d^{{\tau}}\}$ for ${\tau} \in \mathcal{T}_{meta-test}$, domain operators $O$, environment $\mathcal{E}$, maximum iterations $\alpha$
\For {$d^{{\tau}}\in \mathcal{D}_{meta-test}$}
    \State $s_0 \gets Initialize(\mathcal{E})$ \Comment initial continuous state
    \State $\hat{S}_0 \gets SGN(s_0)$, $\hat{S}_G \gets SGN(d^{{\tau}})$
    \While {iteration $ < \alpha$}
        \State $\Pi \gets CP(\hat{S}_0, \hat{S}_G, O)$ \Comment run planner
        \State $s_c \gets ExecActions(\mathcal{E}, \Pi)$ \Comment execute plan
        \State $\hat{S}_0 \gets SGN(s_c)$ \Comment update initial symbolic state
    \EndWhile
\EndFor
\end{algorithmic}
\end{algorithm}

\setlength{\textfloatsep}{4mm}
\begin{figure*}[t]
  \centering
  \includegraphics[width=0.9\linewidth]{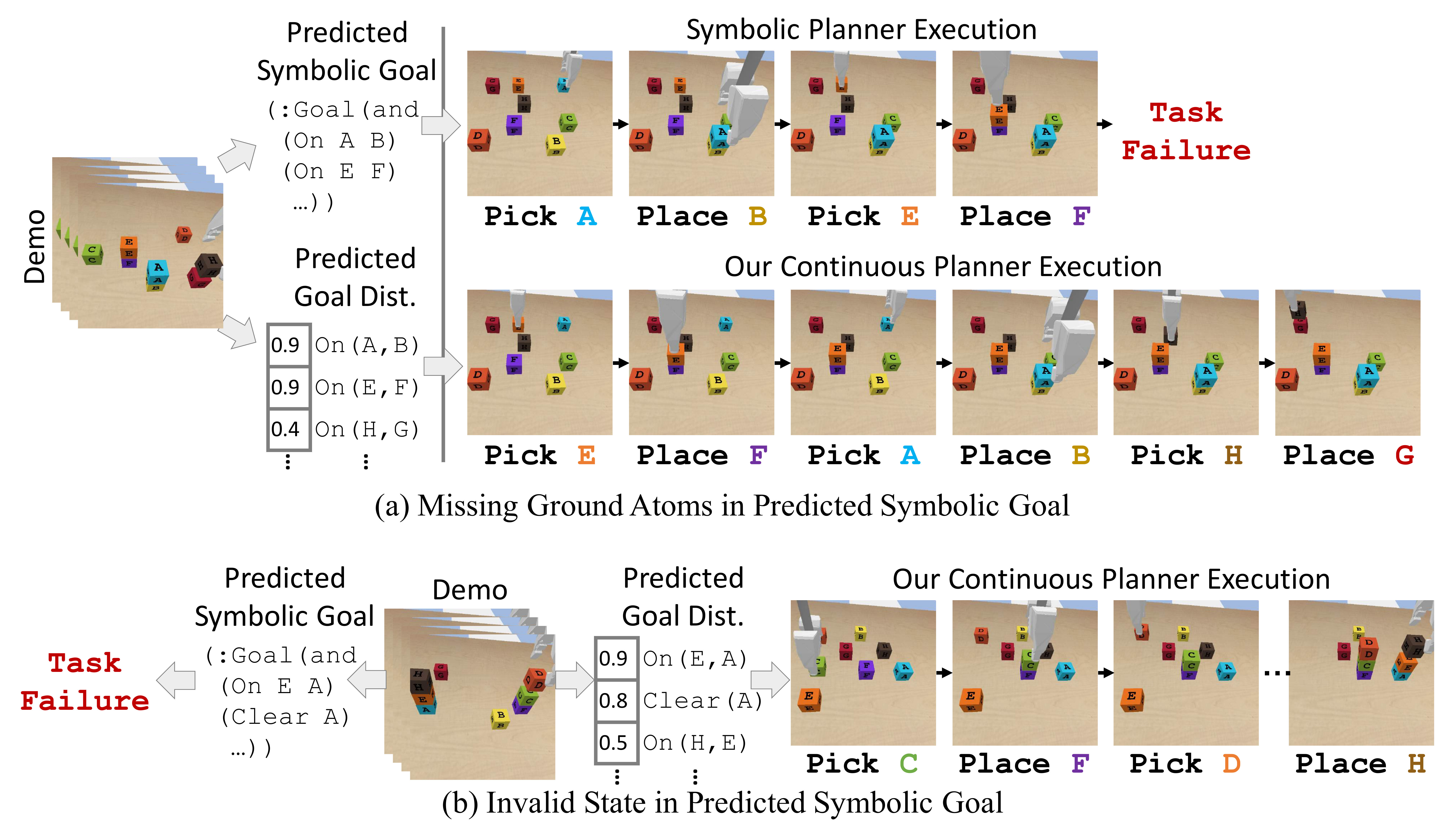}
  \vspace{-2mm}
  \caption{Block stacking qualitative results. (a) The predicted discrete symbolic goal failed to capture that \texttt{On(H,G)} and does not complete the task. On the other hand, the proposed continuous planner can still complete the task by aiming to match the predicted goal distribution. (b) The predicted symbolic goal is invalid as it includes both \texttt{On(E,A)} and \texttt{Clear(A)}. In this case, it is impossible to reach this goal state with the symbolic planner. On the other hand, our continuous planner can still generate sequences of actions guided by \texttt{On(E,A)} to best match the predicted goal distribution.
  }
  \vspace{-2mm}
  \label{fig:qual}
\end{figure*}

\vspace{1mm}
\noindent\textbf{Goal Satisfaction.} Similarly, the goal satisfaction condition is no longer defined by the presence of the ground atom in the symbolic state because neither the goal nor the current state is symbolic. Instead, we have the current distribution over symbolic state, and the distribution of goal state. The objective of our search is thus to match the two distributions.

With these continuous relaxations, we have defined all the operations (i) to (v) on the symbolic states in terms of the distribution over states. In addition, our derivation is efficient as it only iteratively operates on $P_{Z(s)}(g)$ that can directly be mapped to the SGN outputs using \eqnref{sgn}. In this case, our Continuous Planner can directly run the search to match the goal distribution on the outputs of SGN.  Note that our Continuous Planner would have the exact same behavior as the symbolic planner if the distribution concentrates on a single state. Therefore, Continuous Planner is generalizing the symbolic planner to handle distribution of states.

\subsection{Learning and Inference}
\label{sec:learning}

\noindent\textbf{Learning.} Our Continuous Planner is a continuous relaxation of the symbolic planner and does not require training. Therefore, we only have to train on SGN. We learn \eqnref{sgn} with full supervision. The procedure is summarized in \algref{training} and \figref{sys_fig}(a). $SymbolicState(\cdot)$ computes the aligned symbolic state $S^\tau$ for each $d^\tau$ based on the action annotation $a^\tau$ used in previous works~\cite{huang2019neural,xu2018neural}.

\noindent\textbf{Inference.} As defined in \eqnref{decomp}, our model can act as a closed-loop policy conditioned on $d^{\tau}$. The procedure is summarized in \algref{testing} and \figref{sys_fig}(b). As the predictions $\hat{S}_G$ and $\hat{S}_0$ are simply distributions over states, it is possible that the model fails to reach the goal after executing the actions in $\Pi$. In this case, we update the initial symbolic state $\hat{S_0}=SGN(s_c)$ and re-plan.

\section{EXPERIMENTS}
\label{sec:experiments}

Our goal is to produce a closed-loop policy to solve a previously unseen task based on a single demonstration. The key insight is to formulate it as a planning problem along with the symbol grounding problem. We propose Continuous Planner  that directly operates on the outputs of the Symbol Grounding Networks.
Our experiments aim to answer the following questions: (1) How does our formulation of one-shot imitation learning as a planning problem compare to alternative policy representations? (2) How well does our continuous relaxation of Continuous Planner handle the outputs of our Symbol Grounding Networks? We answer these questions by evaluating our method on two task domains: Block Stacking and Object Sorting in BulletPhysics~\cite{huang2019neural}. We compare the proposed framework with alternative formulations of one-shot imitation learning and evaluate the importance of our continuous relaxation.

\vspace{1mm}
\noindent\textbf{Implementation Details.}
We use the object poses as continuous input states to our SGN. Both the object modules and the predicate modules of the SGN are 2-layer perceptrons with 128 hidden units in each layer. For all the planner based methods including our Continuous Planner, we use a simple forward planning algorithm for a fair comparison.

\subsection{Baselines for Comparison}

We compare the following baselines for the experiments:

\vspace{1mm}
\noindent\textbf{Neural Task Graph Networks (NTG)~\cite{huang2019neural}.} NTG is the closest to our work among the deep learning-based one-shot imitation learning  approaches~\cite{duan2017one,xu2018neural,finn2017one}. In contrast to our planning-based formulation, NTG parameterizes the policy with a graphical structure to modularize the policy. This modularity does improve the data efficiency compared to the ones without~\cite{duan2017one,xu2018neural,finn2017one}, and thus we see NTG as our main comparison among all the previous works. Note that in the graph-based formulation, the policy execution is still coupled with the inter-task generalization in contrast to our decoupling using the planning-based formulation.

\begin{figure}[t]
  \centering
  \includegraphics[width=0.9\linewidth]{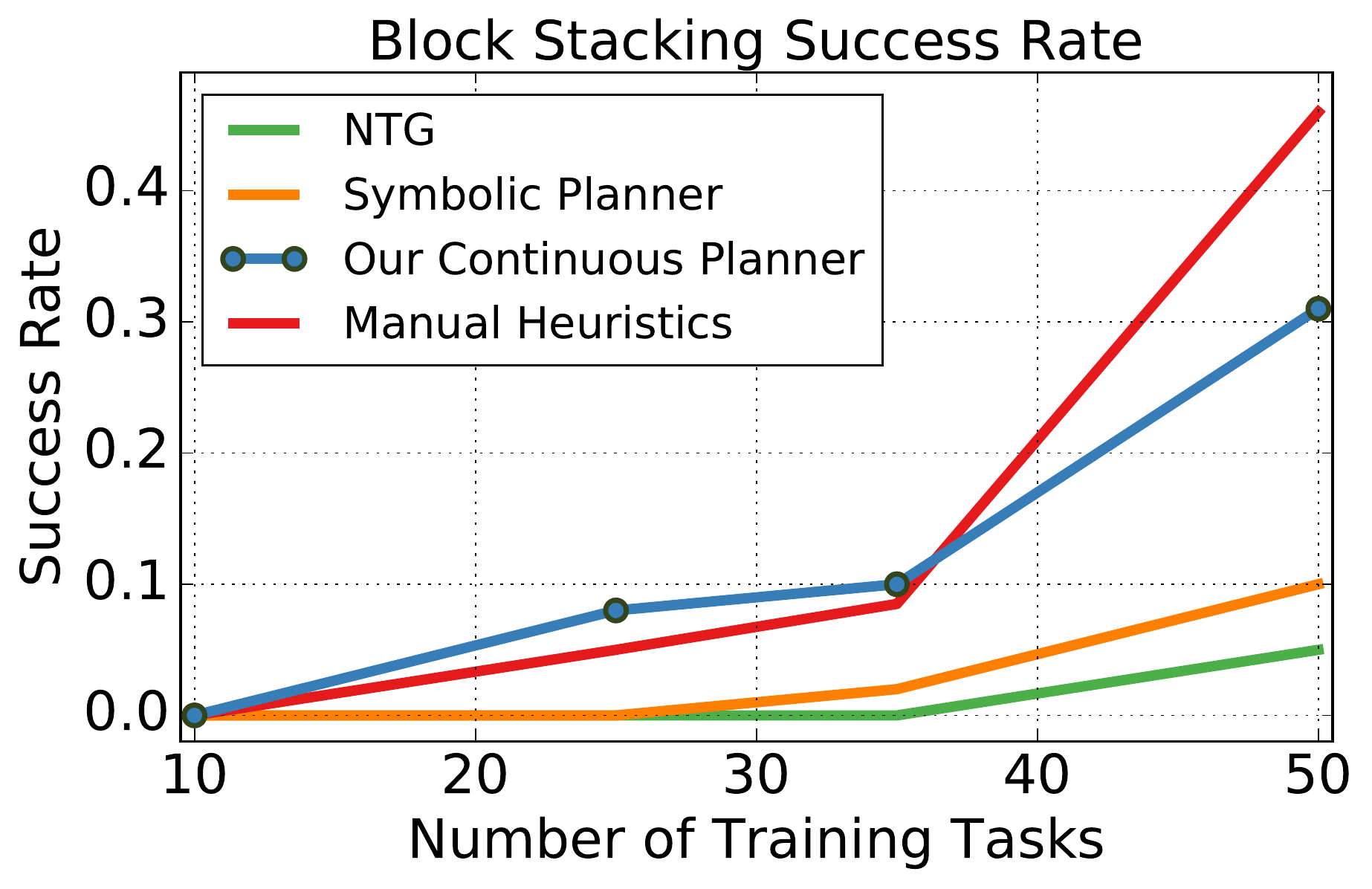}
  \vspace{-2mm}
  \caption{Block Stacking Results. The planning based methods all outperform the policy network of Neural Task Graph Networks~\cite{huang2019neural}. More importantly, our Continuous Planner significantly outperforms the baseline symbolic planner by operating on the distribution over states. This shows the benefits of our continuous relaxation. On the other hand, our Continuous Planner is also comparable to the Manual Heuristics baseline but does not require additional human efforts.
  }
  \vspace{-2mm}
  \label{fig:stacking}
\end{figure}

\vspace{1mm}
\noindent\textbf{Symbolic Planner + Discrete SGN (SP).} As previously discussed, one may discretize the output of the Symbolic Grounding Network and plan with a classical symbolic planner (SP). 
The goal of comparing with SP is to show the effectiveness of our Continuous Planner. We use the same SGN as our Continuous Planner and discretize the outputs by picking the symbolic state with the highest probability from the distribution over states given by the SGN.

\vspace{1mm}
\noindent\textbf{SP + Manual Heuristics.} We have briefly discussed in \secref{CPDDL} that while it is possible to formulate the invalid state checking as a SAT problem, it leads to a computational bottleneck. An alternative way is to manually design rules to identify invalid states. For example, as aforementioned, the state where both \texttt{On(A, B)} and \texttt{Clear(B)} are true is invalid. In addition, one needs to define how to rectify the observed invalid state to a valid one. 
For example, when we see both \texttt{On(A, B)} and \texttt{Clear(B)}, we need to further decide which one to keep to make the state valid. Hence, a baseline is to manually define domain-dependent heuristics to rectify the states. One could also use methods like Markov Logic Networks~\cite{richardson2006markov} with infinite weights on the rules of invalid states for this. The drawback of this baseline is that defining the rules needs extensive domain knowledge, whereas our method is domain-independent.
Again we use the same SGN as SP for a fair comparison.

\subsection{Evaluating Block Stacking}

\noindent\textbf{Experimental Setups.} We follow the setup of Block Stacking tasks in previous works~\cite{huang2019neural,xu2018neural}, where the goal is to stack the set of 8 blocks into a target configuration. The blocks are 5 cm cubes. The final block configuration of the demonstration is used as the goal state. For the Manual Heuristics baseline,
we have three rules for invalid states in this domain. First, if an object is on top of another object, then the bottom object cannot be \texttt{Clear}. Further, a block can only be on top of another block and can only have one other block on top of it.

\begin{figure}[t]
  \centering
  \includegraphics[width=0.9\linewidth]{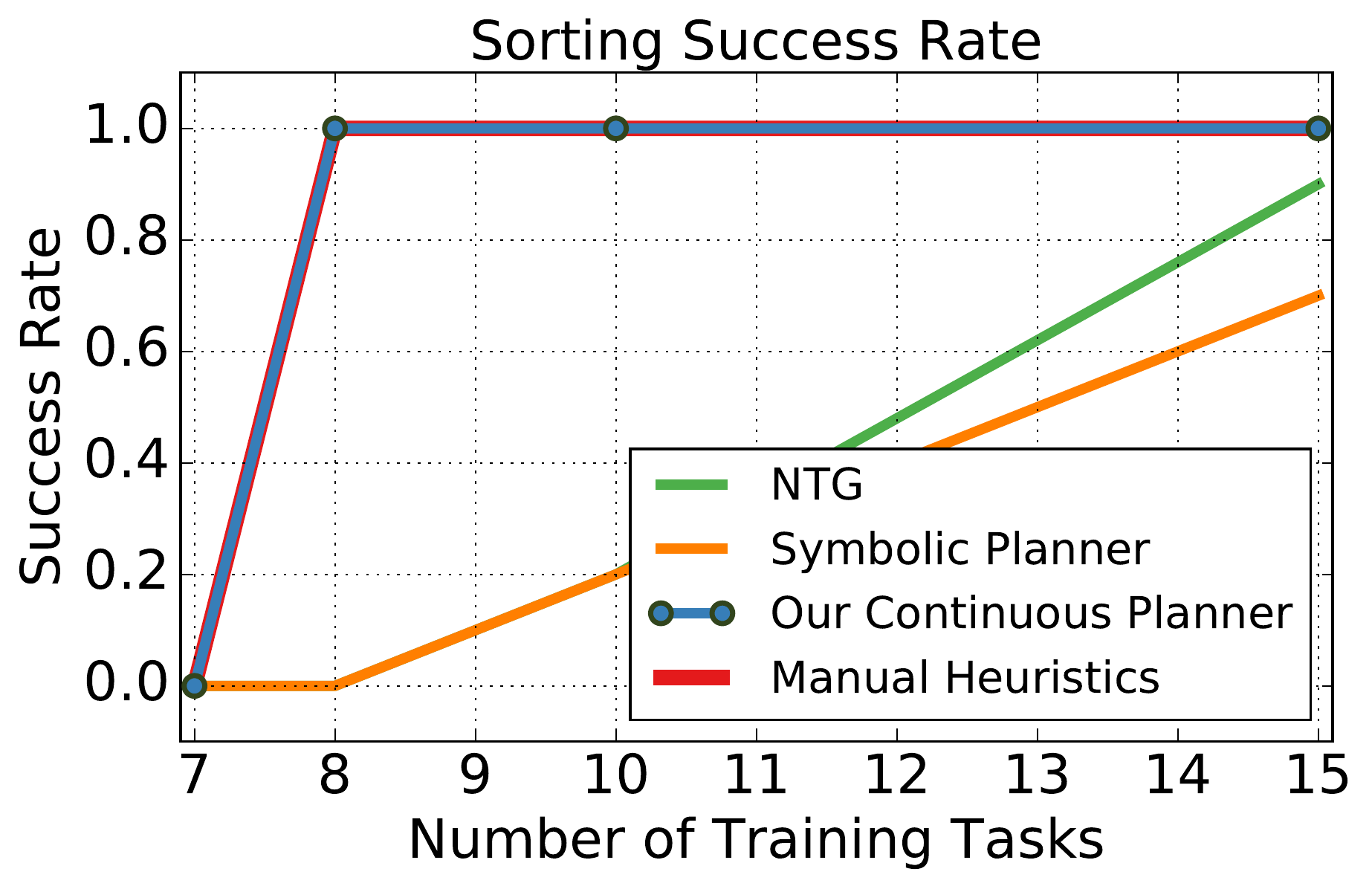}
  \vspace{-2mm}
  \caption{Object Sorting Results. The main challenge is that the model needs to come up with alternative solutions that are different from the demonstration.
  The planning formulation allows us to naturally handle the alternative solutions to the task and quickly learn to complete the task on unseen configurations with only 8 training tasks. This is in contrast to NTG, which learns a task graph generator specifically to address the alternative solutions.
  }
  \vspace{-2mm}
  \label{fig:sorting}
\end{figure}

\vspace{1mm}
\noindent\textbf{Results.} The results are shown in \figref{stacking}. The x-axis shows the number of training tasks used in $\mathcal{T}_{meta-train}$ and the y-axis shows the success rate of the model on tasks in $\mathcal{T}_{meta-test}$ given a single demonstration. All of the planning-based methods outperform the policy networks of NTG~\cite{huang2019neural}, even though the NTG policy is already parameterized by the task graph to improve the data efficiency.
The results show the importance of our formulation of one-shot imitation learning as a planning problem to disentangle the inter-task generalization from the task execution.
Our Continuous Planner is able to directly plan on the probabilistic outputs of SGN and significantly outperform the symbolic planner baseline, in which we are forced to make uninformed discrete decisions based on the SGN outputs that can easily lead to invalid states.
In addition, our Continuous Planner is able to perform comparably to the Manual Heuristics baseline without using any further manually designed rules and heuristics to rectify the invalid states. This demonstrates the domain-independent scalability of our method. 
\figref{qual} compares the proposed Continuous Planner and the symbolic planner baseline. In \figref{qual}(a), the discretized SGN outputs fails to recognize that the goal configuration requires \texttt{On(H, G)}. This prevents the symbolic planner from completing the task. On the other hand, the continuous outputs of SGN are still able to inform our Continuous Planner about the goal \texttt{On(H, G)}. \figref{qual}(b) shows an important case for another task. The original discretized output of SGN is an invalid state, where both \texttt{On(E, A)} and \texttt{Clear(A)} are presented. The symbolic planner fails to reach the goal because of this incorrect state discretization. On the other hand, our Continuous Planner can still operate on this distribution over states to match the goal distribution and successfully complete the task.

\subsection{Evaluating Object Sorting}

\noindent\textbf{Experimental Setups.} The goal of the Object Sorting domain is to move the objects scattered on the tabletop to the corresponding containers shown in the demonstrations. We use four object categories and four containers. We consider the challenging setting of Huang \etal~\cite{huang2019neural}, where the task is initialized in such a way that it requires alternative solutions to a task that are distinct from the demonstration. Similarly, the rules for our Manual Heuristics is that an object cannot be at different locations at the same time. We use this as a hard constraint and maximize the total probability.

\vspace{1mm}
\noindent\textbf{Results.} As shown in \figref{sorting}, the proposed Continuous Planner quickly achieves the perfect performance with only 8 training tasks, without any additional manual rules and heuristics to rectify the outputs of SGN. On the other hand, both of the symbolic planner and the NTG baselines are still unable to achieve 100\% success rate with 15 training tasks. The main challenge is that the model has to infer alternative solutions to the Object Sorting task that are not observed from the demonstration. In order to address the challenge, the NTG baseline specifically designed a task graph generation model to complete the task graph from the demonstration, which enables NTG to outperform the symbolic planner baseline. Compared to NTG, our Continuous Planner approach formulates one-shot imitation as a planning problem and is thus capable of completing the tasks with alternative solutions. This leads to better generalization not only between tasks but also between the alternative solutions to complete the same task.

\section{CONCLUSION}

We presented a new formulation of one-shot imitation learning as a planning problem. This disentangles the inter-task generalization from the policy execution and leads to better data efficiency. The key challenge is that the Symbol Grounding Networks to connect the planner with the continuous input state can be unreliable without sufficient training data. This introduces uncertainty in the state representation. We address this by replacing the set-theoretic representation in the symbolic planner with the probabilistic symbols and generalize the planner to operates on the distribution over states. This allows us to derive the continuous relaxation of the symbolic planner, which significantly improves the performance over the symbolic planner baseline by handling the symbol uncertainty. We show that the resulting Continuous Planner is able to outperform state-of-the-art one-shot imitation learning approaches on two challenging task domains. This shows the importance of both our planning formulation and continuous relaxation.

\vspace{2mm}
\noindent\textbf{Acknowledgements.} 
Toyota Research Institute (``TRI'')  provided funds to assist the authors with their research but this article solely reflects the opinions and conclusions of its authors and not TRI or any other Toyota entity.

\renewcommand*{\bibfont}{\footnotesize}
\begin{flushright}
\printbibliography 
\end{flushright}

\end{document}